\documentclass[10pt,twocolumn,letterpaper]{article}

\usepackage{cvpr}
\usepackage{times}
\usepackage{epsfig}
\usepackage{graphicx}
\usepackage{amsmath}
\usepackage{amssymb}

\usepackage[breaklinks=true,bookmarks=false]{hyperref}

\cvprfinalcopy 

\def\cvprPaperID{****} 

\begin{document}

\title{Pooling Pyramid Network for Object Detection}

\author{
Pengchong Jin
\hspace*{32pt}
Vivek Rathod
\hspace*{32pt}
Xiangxin Zhu
\\
Google AI Perception\\
{\tt\small \{pengchong, rathodv, xiangxin\}@google.com}
}

\maketitle

\begin{abstract}
  We'd like to share a simple tweak of the Single Shot Multibox Detector
  (SSD) family of detectors,
  which is effective in reducing model size while
  maintaining the same quality. We share box predictors
  across all scales, and replace convolution between
  scales with max pooling. This has two advantages over vanilla
  SSD: (1) it avoids score miscalibration across scales; (2)
  the shared predictor sees the training data over all
  scales. Since we reduce the number of predictors to one, and
  trim all convolutions between them, model size is
  significantly smaller. We empirically show that these changes
  do not hurt model quality compared to vanilla SSD.

\end{abstract}

\section{Introduction}
SSD detectors~\cite{liu2016ssd,lin2017focal} have
been popular as they run fast, are simple to implement
and easily portable to different types of hardware.

Most SSD detectors have several feature maps
representing different scales, each of which uses its own
predictor to produce boxes and class scores.
In practice, especially when the data distribution is skewed
over scales, this design is problematic. Imagine a
dataset with many large objects and very few small ones.
The predictors from small scale feature maps will be
wasted as they rarely see any positives. This data imbalance
could also result in score miscalibration across scales even
for the same class. Another issue with this design is that
each predictor only sees the objects at its own scale. This
partition will divide the already small dataset
into even smaller sets. If we believe that object
appearance is scale invariant, it will be more efficient
if all the predictors see all of the data.

We propose simple changes to vanilla SSD: use the same predictor
for all scales. In order for the predictor to work in the
same feature space, we replace convolutions between
feature maps with max pooling.







\section{Pooling Pyramid Network (PPN)}
\begin{figure*}[t]
\begin{center}
\includegraphics[width=1.0\linewidth]{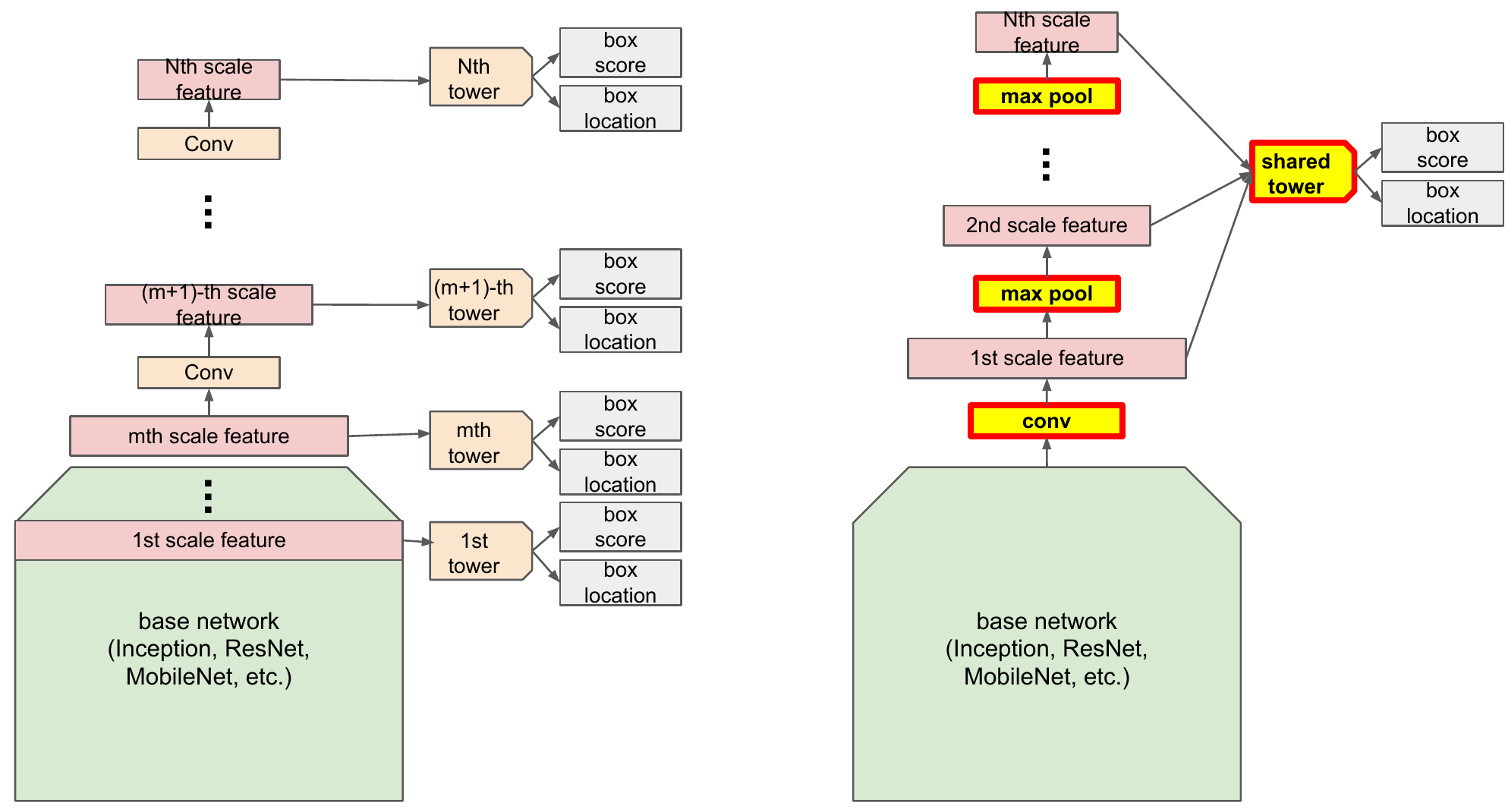}
\end{center}
\caption{
Architecture comparison between the Pooling Pyramid Network (PPN)
and vanilla SSD. Left: vanilla SSD, Right: PPN.
Note that the changes in PPN are highlighted:
(1) using max pool to build the feature pyramid,
(2) using shared convolutional predictors for box classification and regression.
}
\label{fig:ppn}
\end{figure*}

The proposed model, \textit{Pooling Pyramid Network (PPN)},
is a single-stage convolutional object detector, very
similar to vanilla SSD with simple changes.  The prediction head is
designed to be light-weight, fast to run, while maintaining
comparable detection accuracy with vanilla SSD.
The network architecture is illustrated in
Figure~\ref{fig:ppn}.  There are two major changes to
vanilla SSD~\cite{liu2016ssd}: (1) the box predictor is
shared across feature maps with different scales; (2) the
convolutions between feature maps are replaced with max
pooling operations.  In the following sections, we will
discuss the rationale behind these changes and their effects.

\subsection{Shared Box Predictor}
Vanilla SSD uses independent box predictors for feature maps at different scales.
One problem is miscalibration of the prediction scores across different scales.

Since each box predictor is trained independently using only
a portion of the groundtruth boxes that it is assigned to,
different box predictors could see very different number of
positive and negative examples during training.  This
implicit data imbalance causes the problem that scores
from different predictors fall in vastly different ranges,
which makes them incomparable and difficult to use in
subsequent score-based postprocessing steps such as non maximum
suppression.  We design PPN with a shared box predictor
across feature maps of different scales.  As a result, the
box predictor sees all of the training data even when
there is an imbalance in groundtruth box scales. This reduces
the effect of miscalibration and unstable prediction scores.

One could argue that having a separate box predictor for each
scale increases the total capacity, and allows each
predictor to focus on its specific scale. However, we think
that this may not be necessary as objects are mostly scale
invariant.

\subsection{Max Pooling Pyramid}

Our goal is to build a multi-scale feature pyramid
structure, from which we can make predictions using the
shared box predictor.  We achieve this by shrinking a
base feature map from the backbone network several times
using a series of max pooling operations.  This is different
from vanilla SSD where feature maps are built by extracting layers
from a backbone network and shrinking them using additional
convolutions, and FPN where feature maps are built by a
top-down pathway with skip connections.  We choose max
pooling mainly for two reasons.  First, using the pooling
operations ensures feature maps with different scales live
in the same embedding space, which makes training the shared
box predictor more effective.  In addition, since max
pooling does not require any additions and multiplications,
it is very fast to compute during inference,
making it suitable for many latency sensitive applications.


\begin{table*}[t]
\begin{center}
\begin{tabular}{l|c|c|c|c}
Model & mAP & inference FLOPs & number of parameters & GPU inference time\\
\hline
\hline
MobileNet SSD & 20.8 & 2.48B & 6.83M & 27ms \\
\hline
MobileNet PPN & 20.3 & 2.35B & 2.18M & 26ms \\
\end{tabular}
\end{center}
\caption{COCO detection: MobileNet SSD vs MobileNet PPN}
\label{comparison}
\end{table*}

\subsection{Overall Architecture}

The final network architecture of our Pooling Pyramid
Network (PPN) detector is illustrated in
Figure~\ref{fig:ppn}.  Followed by the backbone network, an
optional $1\times 1$ convolution is used to transform the features
from the backbone network to a space with desired
dimensions.  We then apply a series of stride-2 max pooling
operations to shrink the feature map down to $1\times 1$.  A shared
box predictor is applied to feature maps of different scales
in order to produce classification scores and location
offsets of box predictions.  We add one additional shared
convolution in the box predictor after pooling operations to
prepare the feature to be used for predictions.

\section{Experiments}

We run experiments on the COCO~\cite{lin2014coco} detection dataset
and compare the performance of PPN with vanilla SSD.
We use MobileNet v1~\cite{howard2017mobilenet} as the backbone network
and set the input resolution to be $300\times 300$.
Both models use the standard implementation of MobileNet-v1 SSD in
the Tensorflow Object Detection API~\cite{huang2017gmi}.
For PPN,
we extract the layer \textit{Conv2d\_11\_pointwise} as the base feature map,
from which we build 6 pooled feature maps that are of sizes
$19\times 19$,
$10\times 10$,
$5\times 5$,
$3\times 3$,
$2\times 2$, and
$1\times 1$.
A shared $1\times 1$ depth 512 convolution is applied before the box classifier and location regressor.
We use the same anchor design as SSD,
smooth $l_{1}$ loss for box regression,
and focal loss with $\alpha=0.25$ and $\gamma=2$ for box classification~\cite{lin2017focal}.
Our implementation is based on the Tensorflow Object Detection API
and is publicly available under Tensorflow's Github repository.

Both SSD and PPN models are initialized using a MobileNet-v1 checkpoint
that is pre-trained on ImageNet, and
both of them are trained and tested on the splits described in~\cite{huang2017gmi}.
We leverage TPUs~\cite{Jouppi2017tpu} for fast training.
We perform the model benchmark using an Nvidia GeForce GTX TITAN X card.
Table \ref{comparison} shows the comparison between SSD and PPN.
PPN achieves similar mAP (20.3 vs 20.8),
comparable FLOPs and inference time,
but is 3x smaller in model size.

{\small
\bibliographystyle{ieee}
\bibliography{refs}

\begin{thebibliography}{1}\itemsep=-1pt

\bibitem{howard2017mobilenet}
A.~G. Howard, M.~Zhu, B.~Chen, D.~Kalenichenko, W.~Wang, T.~Weyand,
  M.~Andreetto, and H.~Adam.
\newblock Mobilenets: Efficient convolutional neural networks for mobile vision
  applications.
\newblock {\em CoRR}, abs/1704.04861, 2017.

\bibitem{huang2017gmi}
J.~Huang, V.~Rathod, C.~Sun, M.~Zhu, A.~Korattikara, A.~Fathi, I.~Fischer,
  Z.~Wojna, Y.~Song, S.~Guadarrama, and K.~Murphy.
\newblock Speed/accuracy trade-offs for modern convolutional object detectors.
\newblock In {\em {CVPR}}, 2017.

\bibitem{Jouppi2017tpu}
N.~P. Jouppi, C.~Young, N.~Patil, D.~Patterson, G.~Agrawal, R.~Bajwa, S.~Bates,
  S.~Bhatia, N.~Boden, A.~Borchers, R.~Boyle, P.-l. Cantin, C.~Chao, C.~Clark,
  J.~Coriell, M.~Daley, M.~Dau, J.~Dean, B.~Gelb, T.~V. Ghaemmaghami,
  R.~Gottipati, W.~Gulland, R.~Hagmann, C.~R. Ho, D.~Hogberg, J.~Hu, R.~Hundt,
  D.~Hurt, J.~Ibarz, A.~Jaffey, A.~Jaworski, A.~Kaplan, H.~Khaitan,
  D.~Killebrew, A.~Koch, N.~Kumar, S.~Lacy, J.~Laudon, J.~Law, D.~Le, C.~Leary,
  Z.~Liu, K.~Lucke, A.~Lundin, G.~MacKean, A.~Maggiore, M.~Mahony, K.~Miller,
  R.~Nagarajan, R.~Narayanaswami, R.~Ni, K.~Nix, T.~Norrie, M.~Omernick,
  N.~Penukonda, A.~Phelps, J.~Ross, M.~Ross, A.~Salek, E.~Samadiani, C.~Severn,
  G.~Sizikov, M.~Snelham, J.~Souter, D.~Steinberg, A.~Swing, M.~Tan,
  G.~Thorson, B.~Tian, H.~Toma, E.~Tuttle, V.~Vasudevan, R.~Walter, W.~Wang,
  E.~Wilcox, and D.~H. Yoon.
\newblock In-datacenter performance analysis of a tensor processing unit.
\newblock In {\em Proceedings of the 44th Annual International Symposium on
  Computer Architecture}, ISCA '17, pages 1--12, New York, NY, USA, 2017. ACM.

\bibitem{lin2017focal}
T.-Y. Lin, P.~Goyal, R.~Girshick, K.~He, and P.~Doll\'{a}r.
\newblock Focal loss for dense object detection.
\newblock In {\em {ICCV}}, 2017.

\bibitem{lin2014coco}
T.-Y. Lin, M.~Maire, S.~Belongie, J.~Hays, P.~Perorna, D.~Damanan,
  P.~Doll\'{a}r, and C.~L. Zitnick.
\newblock Microsoft {COCO}: Common objects in context.
\newblock 2014.

\bibitem{liu2016ssd}
W.~Liu, D.~Anguelov, D.~Erhan, C.~Szegedy, S.~Reed, C.-Y. Fu, and A.~C. Berg.
\newblock {SSD}: Single shot multibox detector.
\newblock In {\em {ECCV}}, 2016.

\end{thebibliography}
}

\end{document}